\documentclass[lettersize,journal]{IEEEtran}
\usepackage{amsmath,amsfonts}
\usepackage{algorithm}
\usepackage{array}
\usepackage{subfigure}
\usepackage{textcomp}
\usepackage{stfloats}
\usepackage{url}
\usepackage{verbatim}
\usepackage{graphicx}
\usepackage{cite}
\usepackage{xcolor}
\usepackage{bm}
\usepackage{graphicx}
\usepackage{amssymb}
\usepackage{amsmath}
\usepackage[colorlinks=true, citecolor=green, linkcolor=red]{hyperref}  
\usepackage{algorithm}
\usepackage[noend]{algpseudocode}
\begin{document}

\title{CoDiff: Robust Collaborative 3D object deteciton via Diffusion-Based Feature Realignment}

\author{Zhe Huang, Shuo Wang, Yongcai Wang*, Deying Li, Zhaoxin Fan, Lei Wang
    
\thanks{
Zhe Huang is with the School of future Transportation, Chang'an University, Xi'an, China (e-mail: huangzhe21@chd.edu.cn) 

Shuo Wang, Yongcai Wang and Deying Li are with the School of information, Renmin University of China, Beijing, China, 100872 (e-mail:shuowang18@ruc.edu.cn; ycw@ruc.edu.cn; deyingli@ruc.edu.cn) 

Zhaoxin Fan is with the Hangzhou International Innovation Institute, Beihang University, Beijing, China (e-mail: zhaoxinf@buaa.edu.cn)

Lei Wang is with the  University of Wollongong, NSW, Australia (e-mail: lei\_wang@uow.edu.au)

}

}


\markboth{Journal of \LaTeX\ Class Files,~Vol.~14, No.~8, August~2021}%
{Shell \MakeLowercase{\textit{et al.}}: A Sample Article Using IEEEtran.cls for IEEE Journals}


\maketitle

\begin{abstract}
Collaborative 3D object detection holds significant importance in the field of autonomous driving, as it greatly enhances the perception capabilities of each individual agent by facilitating information exchange among multiple agents. 
Most collaborative perception systems rely on an idealized noise-free assumption, ignoring delays and pose inconsistencies during interaction. 
However, real-world deployment inevitably introduces noise, causing severe feature misalignments and performance degradation. 
Diffusion models naturally have the ability to denoise noisy samples to the ideal data,
which motivates us to  explore the use of diffusion models to address feature misalignment in multiagent systems caused by noise.
In this work, we propose CoDiff, a novel robust collaborative perception framework that leverages the potential of diffusion models to generate more comprehensive and clearer feature representations.
Specifically, CoDiff consists of two core modules: a Pose Calibration Module (PCM), which establishes associations between objects detected by different agents and optimizes their relative poses by minimizing alignment errors; and a Time Calibration Module (TCM), which leverages single-agent features as conditions to remove delay-induced noise and generate clearer and more comprehensive collaborative representations.
Experimental study on both simulated and real-world datasets demonstrates that the proposed framework CoDiff consistently outperforms existing relevant methods in terms of the collaborative object detection performance and exhibits highly desired robustness when the pose and delay information of agents is with high-level noise.  Ablation studies further validate the effectiveness of key components and design choices within the framework.
\end{abstract}


\section{Introduction}
\IEEEPARstart{3}{D}  object detection \cite{pointpillars, 3d, pointnet++, votenet++,TMM1,TMM2} is a fundamental task in autonomous driving. 
It has gained significant interest in domains including drones, robotics, and the metaverse \cite{I1,I2,D1,D2,TMM3,TMM4}. However, 3D object detection with a single agent has inherent limitations, such as occlusion and missing detection of distant objects 
\cite{qi2017pointnet,huang,huang2,TMM5}.
Recently, researchers have addressed these limitations by enabling multiple agents to share complementary perception information through communication, leading to more comprehensive and accurate detection.
Cooperative 3D object detection \cite{HEAL, where2comm, roco,SycNet} is an important field that has made significant progress  in terms of high-quality datasets and methods. However, it also faces many challenges, including time delays \cite{CoBEVFlow}, pose inconsistencies \cite{roco}, communication bandwidth limitations \cite{where2comm}, and adversarial attacks \cite{adversarial}.

Existing methods have considered various methods to address these challenges. CoAlign \cite{CoAlign} introduces a novel agent-object pose graph modeling approach to improve pose consistency between collaborative agents. RoCo \cite{roco} designs a matching and optimization strategy to correct pose inconsistencies. V2XViT \cite{v2x-vit} incorporates latency as an input for feature compensation, while CoBEVFlow \cite{CoBEVFlow} introduces the concept of feature flow to handle irregular time delays. 
The above methods work well when the noise is minimal and the number of vehicles is limited. However, in crowded and complex scenarios, as noise increases, misalignment in object information transmitted by different agents increases, making it increasingly difficult to align features at the same location. This situation ultimately reduces the accuracy of 3D object detection.

\begin{figure}[t]
\centering
\includegraphics[width=1\columnwidth]{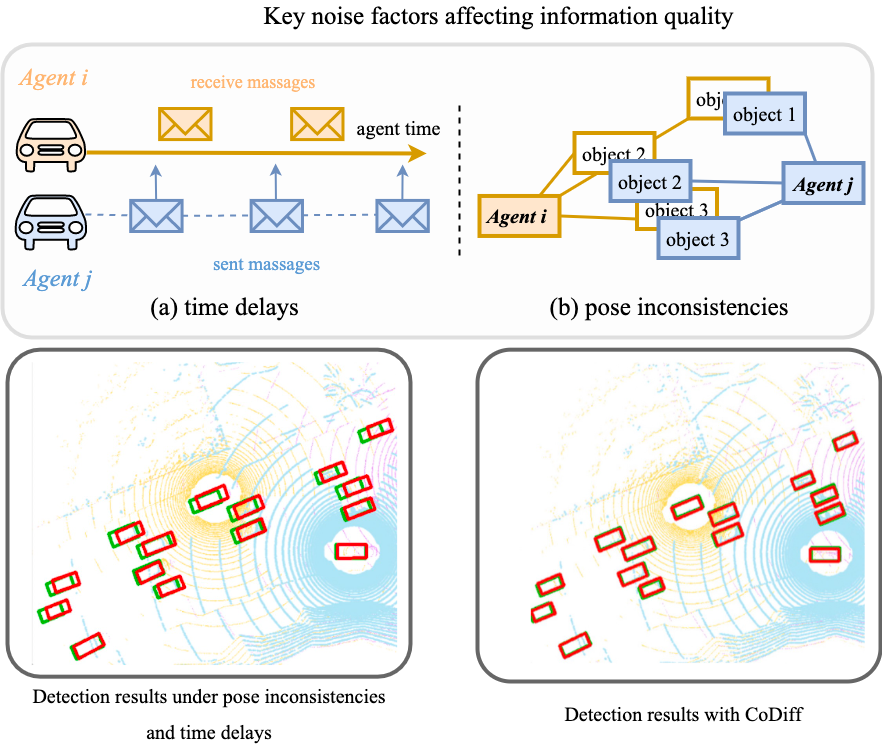} 
\caption{Illustration of  collaborative perception under noise and the perception results w.o./w. CoDiff. Red boxes are detection results and green boxes are the ground truth. }
\label{Fig1}
\end{figure}

To effectively share information, multiple agents need to transmit precisely aligned real-time features to synchronize their data within a consistent timestamp and spatial coordinate system, which is the foundation for maintaining effective collaboration.
However, in real-world scenarios, we observe that the transmitted information may suffer from pose and time misalignment. As illustrated in Fig.\ref{Fig1}, two key factors significantly affect the performance of collaborative perception. The first issue is time delays, as shown in Fig.\ref{Fig1}(a), when agent $j$ sends information to agent $i$, unstable inter-agent communication (e.g. congestion or interruptions) may introduce delays, resulting in misaligned timestamps between agents.
Secondly, the 6-DoF pose estimation conducted by each agent's localization module is inherently imperfect, inevitably leading to relative pose inconsistencies. For instance, in Fig.\ref{Fig1}(b), a noticeable positional deviation between \textit{ object 3} detected by agent $i$ (orange box) and \textit{ object 2} detected by agent $j$ (blue box), which adversely affects subsequent feature fusion. When pose inconsistencies or time delays are large, this misalignment becomes even more pronounced.
Therefore, this work focuses on addressing the inter-agent feature misalignment in 3D object detection caused by time delays and pose inconsistencies, with the goal of mitigating the negative impact of such noise on collaborative detection systems.

\begin{figure}[t]
\centering
\includegraphics[width=1\columnwidth]{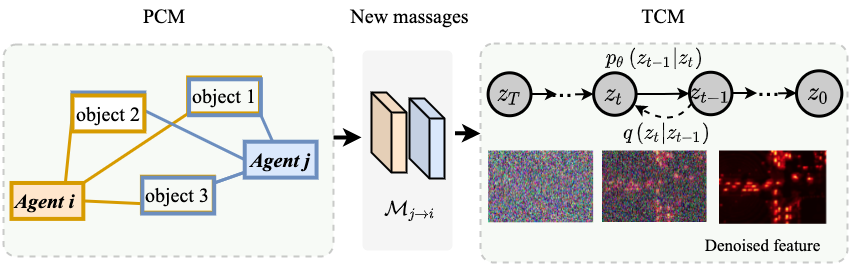} 
\caption{Illustration of the proposed robust 3D multi-agent collaborative perception system with PCM and TCM module, where  $q$ is the diffusion process and $p_{\theta}$ is the reverse process, massage $M_{j\rightarrow i}$ is the condition.}
\label{Fig2}
\end{figure}

To address this limitation, we propose a novel hybrid collaborative framework called CoDiff, which can address pose and time noise in information sharing, mitigate feature misalignment, and ultimately enhance detection accuracy.
In CoDiff, we treat the time delays between multiple agents as a unified noise to be learned and handled, and  employ conditional diffusion models for feature fusion, replacing the existing methods based on regression models and attention mechanisms.
CoDiff primarily consists of two components: 
 a Pose Calibration Module (PCM) and a Time Calibration Module (TCM). 
PCM establishes correspondences among multiple objects and constructs a pose optimization graph to effectively adjust agent pose inconsistencies. TCM first compresses the high-dimensional features of agents into a latent space and then employs a conditional diffusion  model   to progressively synthesize denoised, higher-quality features,   as illustrated in Fig.~\ref{Fig2}.
CoDiff has three main advantages:
i) The pose adjustment method in CoDiff is unsupervised, requiring no ground-truth pose annotations for either agents or objects, and it exhibits strong adaptability to various pose inconsistencies;
ii) Unlike existing methods that design specialized fusion modules, the proposed fusion approach is a generative method based on diffusion models, which fundamentally prevents the introduction of additional noise into the features;
iii) Even in crowded and complex environments,
 CoDiff can accurately align the agents and improve detection accuracy. 

We conducted extensive experiments on both simulation and real-world datasets, including V2XSet\cite{v2x-vit}, OPV2V \cite{opv2v} and DAIR-V2X\cite{dair}. 
Results show that CoDiff consistently achieves the best performance in the task of collaborative 3D object detection with the presence of pose inconsistencies and time delays.  In summary, the main contributions of this work are as follows:

\begin{itemize}  
\item We propose CoDiff, a novel and powerful diffusion-based collaborative perception framework that addresses pose inconsistencies and time delays among multiple agents, alleviates feature misalignment, and improves collaborative detection performance.

\item CoDiff first establishes object correspondences through graph matching, refines their relative poses by minimizing alignment errors, and subsequently employs a diffusion-based generative fusion strategy to mitigate time delays, ultimately yielding clearer and more comprehensive representations.

\item Extensive experiments have shown that CoDiff consistently enhances the accuracy and robustness of 3D object detection, even under challenging conditions with dense traffic and significant noise.
\end{itemize}

\section{Related Work}

\textbf{Collaborative 3D Object Detection.}
Collaborative 3D object detection \cite{syncnet,CoAlign,cobevt,v2x-vit,opv2v} enables multiple agents to share information to overcome the limitations of single-agent perception, and various methods have been proposed to improve its performance and robustness.
In terms of perception performance, the study in the literature \cite{cobevt,v2x-vit,opv2v,where2comm} implemented a transformer architecture to aggregate information from different agents. 
To address communication delays, SyncNet\cite{syncnet} utilizes previous multiframe data to augment present information, implementing feature attention co-symbiotic estimation and temporal modulation approaches.
In terms of robustness, CoBEVFlow\cite{CoBEVFlow} creates a synchrony-robust collaborative system that aligns asynchronous collaboration messages sent by various agents using motion compensation. HEAL\cite{HEAL} smoothly integrates emerging heterogeneous agent types into collaborative perception tasks. CoAlign \cite{CoAlign} uses an agent-object pose graph to address pose inaccuracies, and RoCo \cite{roco} tackles pose inconsistencies with object matching and graph optimization techniques.
However, these methods are generally designed for scenarios with a single type of noise and relatively simple traffic conditions, and their accuracy and robustness are not guaranteed in dense, complex environments. In contrast, CoDiff explicitly accounts for multiple noise sources, including time delays and pose inconsistencies, thereby improving feature fusion quality and ultimately enhancing the accuracy of collaborative 3D object detection.

\textbf{Diffusion Models for Perception Tasks.}
Diffusion model (DM) \cite{D1,D2,D3,D4}, known for their denoising capability, have gained attention in domains such as natural language processing, text transformation, and multimodal data generation, and are increasingly explored in perception tasks. DDPM-Segmentation \cite{DDPMSegmentation} is the initial study that uses diffusion models for semantic segmentation. On the other hand, DiffusionDet \cite{DiffusionDet} considers object detection as a noise-to-box task, aiming to provide accurate object bounding boxes by  denoising randomly generated proposals.  Expanding upon this, Diff3DETR \cite{diff3detr} extends the approach of Diffusion-SS3D by proposing the first diffusion-based DETR framework.
In addition, DifFUSER \cite{diffbev} utilizes the noise reduction capabilities of diffusion models to address noise for multimodal fusion (such as image, text, and video). In order to enable DM training under limited computational resources, Rombach \cite{latent_diffusion} proposed the Latent Diffusion Model (LDM), which can significantly improve training and sampling efficiency without compromising the quality of the denoising diffusion model.
In this work, we are motivated to further explore the potential of employing
the diffusion model to generate a high-quality representation, proposing a diffusion-based multi-agent cooperative perception 3D detection method.


\section{Collaborative Perception and Our proposed method}

\subsection{Collaborative Perception}
It is assumed that there are $N$ agents (i.e., collaborators) in the scene. 
For each agent $i$, let $\mathcal{X}_{i}$ denote its input point cloud 
and $\mathcal{O}_{i}$ denote its corresponding output. 
Furthermore, $\mathcal{M}_{j \rightarrow i}$ represents the collaboration message 
transmitted from agent $j$ to agent $i$.
The goal of collaborative 3D object detection is to accurately classify 
and localize the 3D bounding boxes of objects in the scene 
through the cooperation of multiple agents. 
This process can be formally defined as follows:
\begin{subequations}
\begin{align}
	F_{i}&=\Phi_{Enc} \left( \mathcal{X}_{i}\right),\quad i=1,\cdots,N,  \\
    \mathcal{M}_{j\rightarrow i} &=\Phi_{Proj} \left( \xi_{i}^{t} ,\left( F_{j},\xi_{j}^{t} \right) \right) ,\quad j=1,\cdots ,N;j\neq {i}  \\
 F^{\prime }_{i}&=\Phi_{Agg} \left( F_{i},\{ \mathcal{M}_{j\rightarrow i}\}_{j=1,2,...,N; j\neq{i}} \right), \\
 \mathcal{O}_i&=\Phi_{Dec} \left( F^{\prime }_{i}\right).  
\end{align}
\end{subequations}

Within the collaborative perception framework, each agent $i$ 
extracts features $F_{i}$ from its raw point cloud observations $\mathcal{X}_{i}$ 
using an encoding network, as shown in Step (1a). 
Then, in Step (1b), each agent $j$ employs the projection module 
$\Phi_{Proj}\left( \cdot \right)$ to transform its feature $F_{j}$ 
into the coordinate system of agent $i$ based on $\xi_i$, $\xi_j$, and $t$, 
where $\xi_i$ and $\xi_j$ denote the poses of agents $i$ and $j$, 
and $t$ denotes the timestamp. 
The projected feature is sent as the collaboration message $\mathcal{M}_{j\rightarrow i}$ to agent $i$. 
After receiving all $(N-1)$ messages, agent $i$ aggregates them with its own feature $F_i$ using $\Phi{Agg}(\cdot)$ to produce the fused feature $F'_i$ (Step 1c), which is finally decoded into the perception output $\mathcal{O}_i$ (Step 1d).

\subsection{Issue of feature misalignment}

As shown above, assuming all networks are well-trained, the fused features $F^{\prime }_{i}$ in Step (1c)  depend heavily on the quality of the information $\mathcal{M}_{j\rightarrow i}$ shared between agents. However, the poses estimated by each agent are not perfect in practice, i.e., $\{\xi_{j, j=1,\cdots, N\}}$ have errors, which adversely affects the accuracy of feature projection. Moreover, time delays may cause inconsistencies in the transmitted feature 
$\mathcal{M}_{j \rightarrow i}$, leading to feature misalignment when agent $i$ 
processes the message, which ultimately degrades the performance of 3D object detection.

To address this feature misalignment, existing works have explored different strategies to enhance the quality of fused features across multiple agents, including pose calibration \cite{roco}, time compensation \cite{CoBEVFlow}, and feature fusion techniques \cite{v2x-vit}. However, these methods are designed to handle a single type of error in isolation. In crowded and dynamic environments, agents are simultaneously subject to both pose inaccuracies and communication delays. Under such conditions, this methods are prone to introducing significant misalignment errors during the feature fusion stage. These misalignment errors will in turn degrade localization accuracy and ultimately compromise the overall performance of collaborative perception.


\begin{figure*}[t]
\centering
\includegraphics[width=\textwidth]{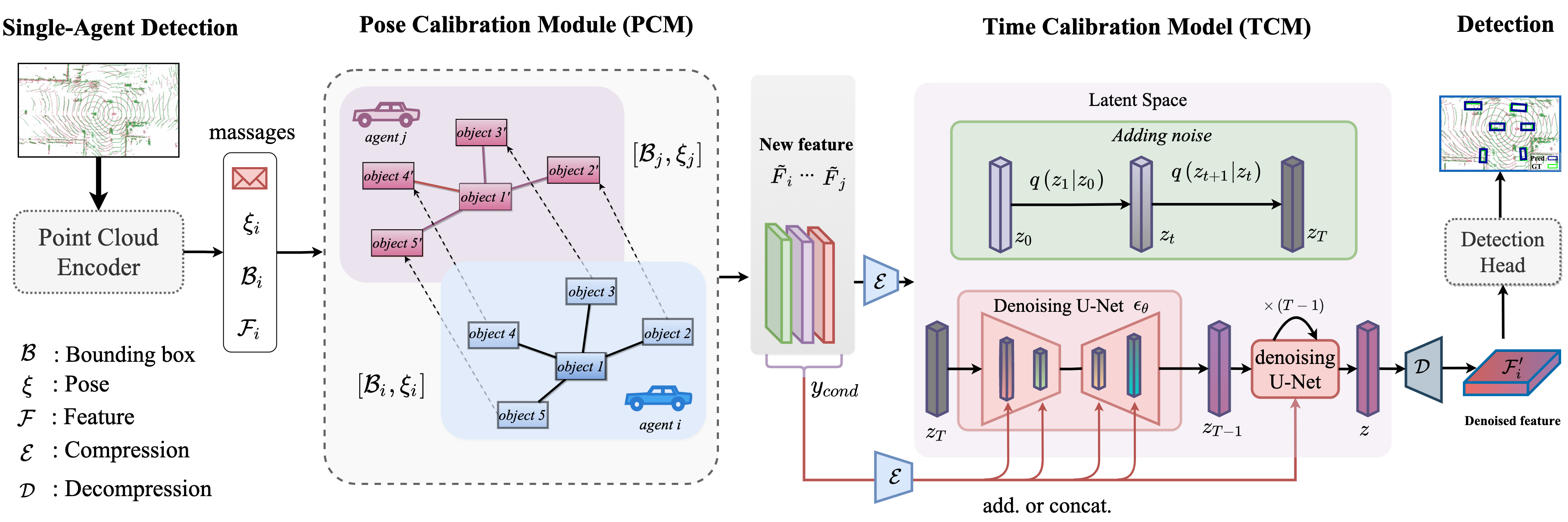} 
\caption{ 
Overview of the CoDiff system. Bounding boxes and poses are transmitted as messages to other agents, where pose calibration is performed through object matching and graph optimization. The calibrated poses are used to transform features into the ego coordinate system, a conditional diffusion model is leveraged to mitigate time delay noise, thereby generating more robust and reliable multi-agent fused features.
}
\label{fig:all}
\end{figure*}

\section{Our Proposed Method}
To improve this situation, we propose CoDiff, a generative model specifically for multi-agent 3D object detection. As illustrated in Fig.~\ref{fig:all}, the overall architecture of CoDiff comprises four major components: a single agent detection module, a pose calibration module (PCM), a  time calibration module (TCM) and a detection head. 
Fig.~\ref{fig:all} shows the overall architecture of CoDiff.
It consists of two key ideas. Firstly, PCM constructs local graphs and exploits graph similarity to establish inter-object correspondences, followed by graph optimization for pose refinement. Secondly, TCM employs a conditional diffusion model to filter time delay noise, thereby generating more reliable multi-agent features.
The fused features are finally fed into the detection head for high-precision 3D object detection.

\begin{figure}[t]
\centering
\includegraphics[width=1\columnwidth]{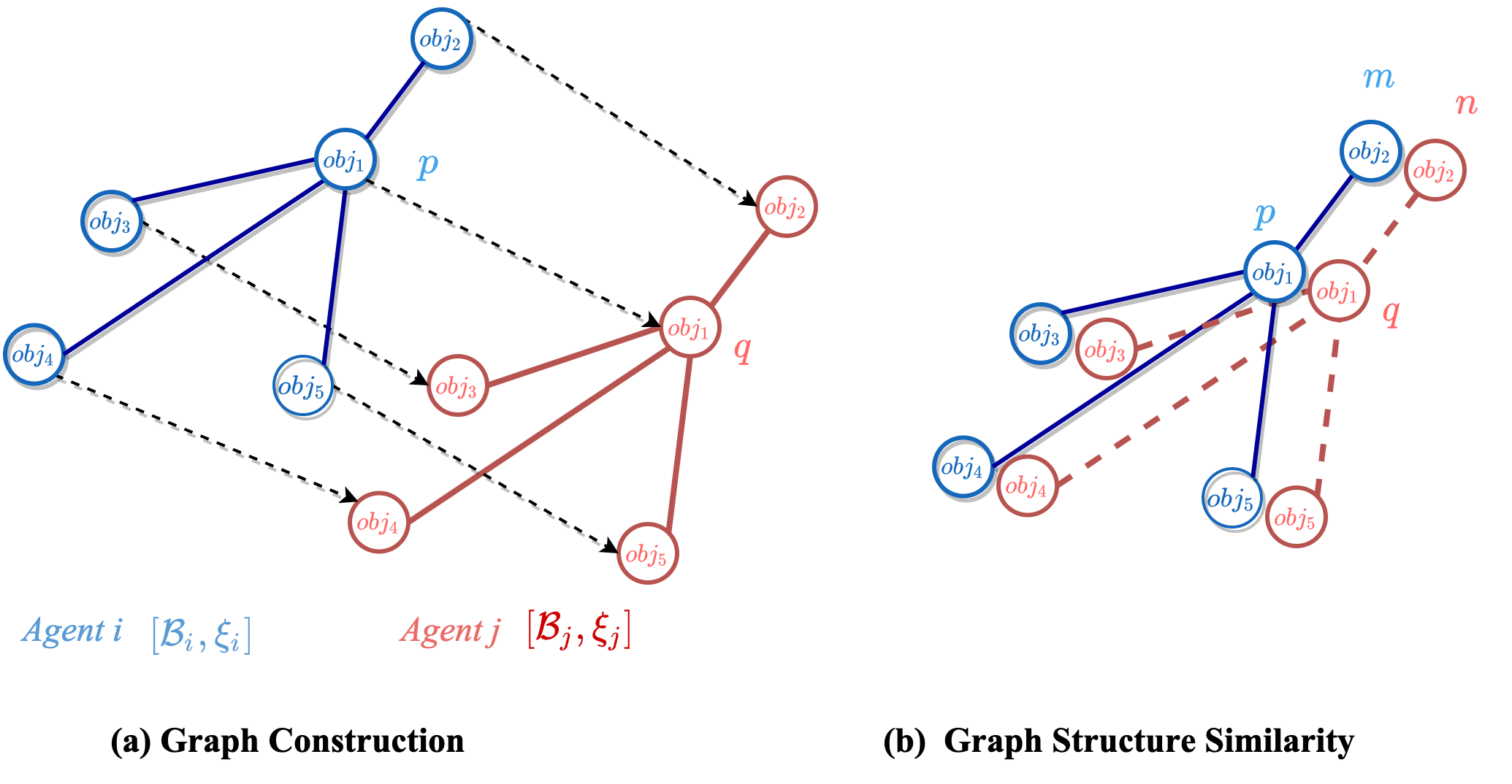} 
\caption{Graph construction and similarity. }
\label{Fig-mo}
\end{figure}

\subsection{Pose Calibration Module}\label{PCM}
The input point cloud is represented as a tensor of size $\left( n\times 4\right)  $, where $n$ indicates the total number of points. 
Each point contains its intensity value along with its 3D coordinates in the global reference frame.
Furthermore, the system estimates the detection uncertainty by computing the variance distributions of both centroid coordinates and heading angles for each bounding box. All agents utilize a unified 3D object detector, denoted as $\Phi_{Enc}$, and every agent can independently and accurately detect objects within its own range.


After completing the initial detection, each agent $j$ transmits a message $\mathcal{M}_{j\rightarrow i} = (F_j, \mathcal{B}_j, \xi_j)$ to the ego-agent $i$, where $F_j $ denotes the feature map, $\mathcal{B}_j$ represents the set of detected bounding boxes, and $\xi_j=\left( x_{j},y_{j},\theta_{j} \right)$ specifies the agent's pose, where $(x_j, y_j)$ is the 2D position and $\theta_j$ the yaw angle relative to the heading angle.
Upon receiving messages from all other agents, the ego-agent $i$ sequentially processes each message to perform graph-based pose calibration. 

The task can be formally defined as follows: given two sets of information $\left\{\mathcal{B}_{i},\xi_{i} \right\}$ and $\left\{\mathcal{B}_{j},\xi_{j} \right\}$, we need to establish correspondences between the two sets of bounding boxes, as shown in Fig. \ref{Fig-mo}(a). To achieve this, we formulate the task as a bipartite graph matching problem \cite{grap-matching1}, where optimal matches are determined through graph similarity measures.
The aforementioned procedure can be formally expressed as:
\begin{subequations}
\begin{align}
\mathcal{A}^*_{ij} &= \{(p,q) \in \mathcal{B}_i \times \mathcal{B}_j \mid S(p,q) \geq \tau_1\} \\
q &= \mathcal{A}_{ij}(p) 
\end{align}
\label{eq:all}
\end{subequations}
Where $p \in \mathcal{B}_i$ and $q \in \mathcal{B}_j$ denote detected objects by agents $i$ and $j$ respectively.
$\mathcal{A}_{ij}$ defines a bijective mapping between the object sets, associating each $p $ with its corresponding $q = \mathcal{A}_{ij}(p) $. 
The matching quality is evaluated through a graph similarity measure $S(p,q)$. A threshold $\tau_1$ is a predefined similarity threshold for reliable correspondences.

\textbf{Graph construction.}
The proposed PCM module is graph-guided.
As illustrated in Fig.~\ref{Fig-mo}(b), for each object $p \in \mathcal{B}_i$, we construct a star-shaped graph $\mathcal{G}_p$ (shown in blue) centered at $p$, where each node represents a feature vector composed of the 3-DoF pose from the object's bounding box. Similarly, for every object $q \in \mathcal{B}_j$, we build a corresponding star-shaped graph $\mathcal{G}_q$ (shown in red). 

\textbf{Initial Matching.} To initiate the matching process, we first transform the bounding boxes \( \mathcal{B}_j \) from the coordinate system of agent \( j \) into the coordinate frame of agent \( i \). Then a distance-based strategy is designed to carry out the initial association.
Specifically, for a pair of candidate graphs \( \mathcal{G}_p \) and \( \mathcal{G}_q \), we compute the spatial distance between their central nodes \( p \) and \( q \), denoted as \( dis(p, q) \). If this distance is below a predefined threshold \( \tau_2 \), the pair \( \mathcal{G}_p \) and \( \mathcal{G}_q \) is considered potentially corresponding and will be included in the matching candidates. It is worth noting that multiple objects from \( \mathcal{B}_j \) may satisfy this distance constraint with respect to node \( p \). In such cases, we select the one with the minimum distance as the initial match, formulated as
\begin{align}
    \mathcal{A} _{ij}(p)=\underset{p\in \mathcal{B}_{i},q\in \mathcal{B}_{j}}{\mathrm{arg}\min} \  \  dis\left( p,q\right)  ;\  where\  dis\left( p,q\right)  \leq \tau_2 
    \label{eqn:tau_2}
\end{align}

\textbf{Graph  Matching.}
Logically, when two detections $p$ and $q$ correspond to the same object, their associated subgraphs $\mathcal{G}_p$ and $\mathcal{G}_q$ should have high similarity.  Taking this into account, we design two similarity metrics, edge similarity and distance similarity, to jointly evaluate the proximity between graphs $\mathcal{G}{p}$ and $\mathcal{G}{q}$, thus determining whether $p$ and $q$ originate from the same object.

\begin{itemize}
    \item \textbf{Edge Similarity.}
In the graph $\mathcal{G}_p$, each edge $e_{pm}$, which connects the central node $p$ to its neighboring node $m$, encodes the relative pose transformation 
$\mathbf{T}_{pm} \in \mathbb{R}^{3\times 3}$. This transformation 
$\mathbf{T}_{pm}$ is a $3\times 3$ matrix computed based on the detection outputs of objects $p$ and $m$. Analogously, in the graph $\mathcal{G}_q$, the edge $e_{qn}$ connecting node $q$ to its neighbor $n$ encodes the relative pose transformation 
$\mathbf{T}_{qn}$.

When $p$ and $q$ correspond to the same  object (i.e.,  $\mathcal{G}_p$ and $\mathcal{G}_q$ exhibit high structural similarity), their edge transformations should satisfy the consistency condition: $\mathbf T_{qn}$ should be consistent with $\mathbf T_{pm}$. We establish the following criterion to assess the consistency of the edge between $e_{pm}$ and $e_{qn}$:
\begin{equation}
l(e_{pm}, e_{qn}) = \exp\left(-\|\mathbf{T}_{pm} (\mathbf{T}_{qn})^{-1} - \mathbf{I}\|_F\right)
\label{eq:edge_consistency}
\end{equation}
where $\|\cdot\|_F$ denotes the Frobenius norm, ${\mathbf I}$ denotes the identity matrix.  The overall edge consistency score   is computed as:
\begin{eqnarray}
\begin{gathered}S_{\text{edge}}(p,q)=\frac{1}{|N_{p}|} \sum_{m\in N_{p}} l(e_{pm},e_{qn})\\ s.t.\  \  \  n={\mathcal A}_{ij}\left( m \right)\end{gathered}
\label{eqn:edge}
\end{eqnarray}
where $N_p$ denotes the leaf nodes in $\mathcal{G}_{p}$ matched with those in $\mathcal{G}_{q}$, $n$ is the counterpart of $m$ from ${\mathcal A}_{ij}$.

\item \textbf{Distance Similarity.} 
Beyond edge similarity, the degree of correspondence between two graphs can be further assessed through distance similarity. The rationale is that when the central nodes $p$ and $q$ represent the same object, the spatial distance of their associated detection boxes should remain minimal, even under noisy measurements or in highly congested scenes. Accordingly, we define the distance similarity between $p$ and $q$ as:
\begin{eqnarray}
S_{dis}\left( p,q\right)  =\exp\left( -dis\left( p, q\right)\right)  
\label{eqn:distance}
\end{eqnarray}
where $dis(.,.)$ computes the Euclidean distance between bounding box centroids, the overall graph similarity is:
\begin{eqnarray}
S(p,q) & = & S_{edge}(q,p) + \lambda S_{dis}(p,q)
\label{eqn:Simila}
\end{eqnarray}
where $\lambda$ is  the hyperparameter for balancing.
\end{itemize}

Finally, the overall similarity $S$ is used to solve the maximization problem in Equation (\ref{eqn:Simila}) to obtain the optimal matching. This is treated as a bipartite graph problem, with one set containing bounding boxes from agent $i$ and the other from agent $j$, using previously defined similarity $S(p,q)$ as edge weights. The Kuhn-Munkres algorithm \cite{KM} efficiently finds the solution, which is applied iteratively across all agents until every pairwise matching is completed.

\begin{figure}[t]
\centering
\includegraphics[width=1\columnwidth]{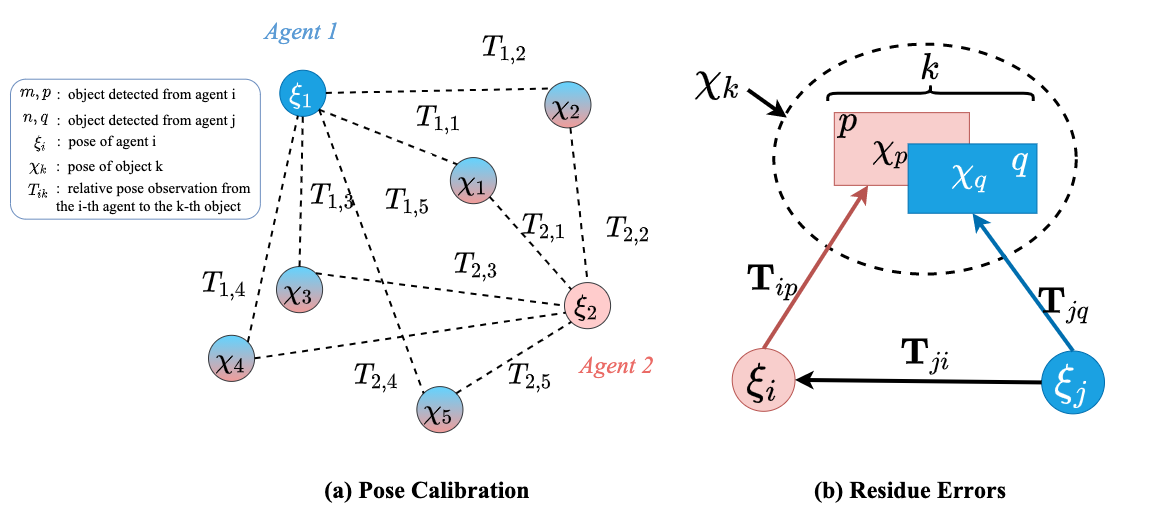} 
\caption{Pose calibration and the residue errors. }
\label{Fig-op}
\end{figure}

\textbf{Pose Calibration.}
After completing the object matching, we obtain a one-to-one correspondence between the elements of the object detection box set $\mathcal B_i$ and $\mathcal B_j$, As illustrated in Figure~\ref{Fig-op}(a).
we construct a pose graph $\mathcal{G}=(\mathcal{V}_{\text{agent}}, \mathcal{V}_{\text{object}}, \mathcal{E})$ for all agents and their detected objects in the scene, using these correspondences to adjust agent poses.   
In this pose graph, $\mathcal{V}_{\text{agent}}$ represents all agents $N$, $\mathcal{V}_{\text{object}}$ contains all detected objects, and $\mathcal{E}$ encodes detection relationships between agents and objects. Each node is associated with a pose. The pose of the agent $j$  is denoted as $\xi_j$, and the pose of the object $k$ is denoted as $\chi_k$.

Consider a matched pair $(p,q)$ where $\mathcal{A}_{ij}(p)=q$. 
Let $\mathbf{T}_{ip}$ denote the relative transformation of object $p$ measured from agent $i$'s coordinate frame, while $\mathbf{T}_{jq}$ represents the relative transformation of object $q$ observed by agent $j$. From agent $i$'s perspective, we establish the relationship:
\begin{equation}
\mathbf{T}_{ip} = \mathbf{E}_i^{-1}\mathbf{X}_p \Rightarrow \mathbf{X}_p = \mathbf{E}_i\mathbf{T}_{ip}
\end{equation}
where $\mathbf{E}_i$ is the transformation matrix constructed from agent $i$'s pose $\xi_i$. Similarly, we derive $\mathbf{X}_q = \mathbf{E}_j\mathbf{T}_{jq}$ for agent $j$'s observations.
The optimization variables in this framework are the agent and object poses $\{\xi_j, \chi_k\}$. The objective is to achieve pose consistency where matched objects converge to identical coordinates:
\begin{equation}
\chi_p = \chi_q = \chi_k
\end{equation}
Figure~\ref{Fig-op}(b) demonstrates our residual error formulation approach. We define the error functions as:
\begin{align}
\mathbf{e}_{ik} &= (\mathbf{E}_i\mathbf{T}_{ip})^{-1}\mathbf{X}_k = \mathbf{T}_{ip}^{-1}\mathbf{E}_i^{-1}\mathbf{X}_k \\
\mathbf{e}_{jk} &= (\mathbf{E}_j\mathbf{T}_{jq})^{-1}\mathbf{X}_k = \mathbf{T}_{jq}^{-1}\mathbf{E}_j^{-1}\mathbf{X}_k \\
\mathbf{e}_{ji} &= \mathbf{T}_{ji}^{-1}\mathbf{E}_j^{-1}\mathbf{E}_i
\end{align}

The overall  optimization objective is formulated as:
{\small
\begin{align}
\left\{ \bm \xi'_j ,\bm \chi'_k \right\}  =\underset{\left\{ \bm \xi_{j} ,\bm \chi_{k} \right\}  }{\arg} \min\sum_{\left( j,k\right)  } {\left( \boldsymbol{e}_{ik}^{T}\mathbf{\Omega }_{ik}\boldsymbol{e}_{ik}\!\!+\!\!\boldsymbol{e}_{jk}^{T}\mathbf{\Omega }_{jk}\boldsymbol{e}_{ik}\!\!+\!\!\boldsymbol{e}_{ij}^{T}\mathbf{\Omega }_{ij}\boldsymbol{e}_{ij} \right)}
\label{eqn:optmodel}
\end{align}
 }
The information matrix $\mathbf{\Omega} = \mathrm{diag}(\sigma_x^{-2}, \sigma_y^{-2}, \sigma_\theta^{-2}) \in \mathbb{R}^{3\times 3}$ encodes the detection uncertainty associated with agent $j$'s bounding box estimates, which are included in the detection outputs. Applying the Levenberg-Marquardt algorithm~\cite{levenberg} to solve~\eqref{eqn:optmodel}, we obtain the optimized poses $\{\bm{\xi}_j'\}_{j=1}^N$ for agents and $\{\bm{\chi}_k'\}_{k=1}^K$ for objects. 
The process converges when the global consistency error is minimized. 


Finally, the corrected relative pose from agent $j$ to agent $i$ is represented as $\xi^{\prime }_{j\rightarrow i} = \xi^{-1}_{i} \circ \xi^{\prime }_{j}$.
Based on this corrected relative pose $\xi^{\prime }_{j\rightarrow i}$, the message $\mathcal{M}_{j\rightarrow i}$ can be aligned accordingly, so that the feature map $\tilde{F}_{j}$ from agent $j$ is transformed into the ego coordinate frame, making it consistent with agent $i$’s own feature $\mathbf{F}_{i}$.

\subsection{Time Calibration Module }
\label{TCM}
 
Upon completing pose calibration, it is necessary to integrate the features from multiple agents to derive the final result. Existing fusion strategies \cite{v2x-vit, where2comm}  rely on autoregressive or attention-based Transformer models. Nevertheless, we observe that in collaborative detection, these methods overlook the fact that time delay is also one of the causes of feature misalignment, thereby leading to fused features with harmful noise.  In TCM, We employed diffusion models for feature fusion, replacing the existing feature fusion methods based on regression models and attention mechanisms. We treat the time delays between multiple agents as a unified noise to be learned and handled. 

Despite their denoising advantage, diffusion models require costly function evaluations during high-resolution feature sampling, leading to significant computational overhead. To address this, we perform sampling in a lower-dimensional latent space, where diffusion models can operate within a perceptually equivalent but computationally efficient representation. This strategy, previously validated in high-resolution image synthesis \cite{latent_diffusion}, motivates the design of our perception compression module.
This autoencoder comprises an encoder and a decoder, focusing on compressing and reconstructing input tensors along the channel dimension. The encoder employs a series of convolutional layers, batch normalization, and ReLU activation functions \cite{relu} to compress high-dimensional input tensors into a lower-dimensional latent space. The decoder, using a similar structure, aims to reconstruct a low-dimensional latent representation back to the original dimensionality. This autoencoder effectively reduces computational overhead while preserving key features of the data.

Precisely, given a feature map $F_i \in \mathbb{R}^{H\times W\times C}$ obtained from PCM, where $H,W,C$ represents height, width and channels, respectively. The encoder ${\mathcal E}$ encodes $F_i$ into a latent representation $z={\mathcal E} \left( F_{i} \right)$ using a compression rate $\tau$, and the decoder $\mathcal D$ reconstructs the feature from the latent space using the same compression rate.
Therefore, the process of compressing the features $F_i$ into the latent space is:
\begin{eqnarray}
\tilde{F}_{i} =\mathcal D\left( z \right) =\mathcal D\left( {\mathcal E}\left( F_{i} \right) \right)
\end{eqnarray}
The perceptual compression loss $\mathcal{L}_{cmp}$ defined as follows
\begin{eqnarray}
\mathcal{L}_{cmp}=D_{KL}\left( F_{i}||\tilde{F}_{i} \right)
\end{eqnarray}
where $D_{KL}(p||q)$ denotes the Kullback-Leibler (KL) \cite{KL} divergence of distribution $p$ from distribution $q$.

With the perceptual compression model, the diffusion model can now be trained in this lower-dimensional, computationally more efficient latent space. 
Our goal is to generate noise-free features $ F^{\prime }_{i}$ (Step (1c)) through this latent space and the decoder $\mathcal D$, which should represent the aggregation of features from multiple agents. 
We can decode the latent representation of features  $ F^{\prime }_{i}$ from the latent space back to the original feature space with a single pass through $\mathcal D$ during training.

\textbf{Conditional Denoising Diffusion Module.}
Diffusion models are generative models inspired by nonequilibrium thermodynamics \cite{non,markov}, designed to learn a data distribution $P\left( x \right)$ by gradually denoising a normally distributed variable. 
The diffusion forward process transforms an initial data sample
$x_0$ into a noisy sample $x_t$ at time $t$ by introducing noise that is determined by the noise variance schedule $\beta_{t}$. For convenience, we denote a series of constants: $\alpha_{t} :=1-\beta_{t} ,\bar{\alpha}_{t} :=\prod_{s=1}^{t} a_{s}$. The forward process to generate a noisy
sample $x_t$ from $x_0$ can be defined by
\begin{eqnarray}
q\left( x_{t}|x_{0} \right) =\mathcal{N}\left( x_{t};\sqrt{\bar{\alpha}} x_{0},\left( 1-\bar{\alpha}_{t} \right) \mathbf{I} \right), 
\end{eqnarray}
where $\mathcal{N}$ denotes a Gaussian distribution and  $\mathbf{I}$ is an identity matrix. The neural network is then trained to
reverse this diffusion process. This network can be interpreted as an equally weighted
sequence of denoising autoencoders $\epsilon_{\theta} \left( x_{t},t \right) ;t=1, ..., T$, which are trained to predict a denoised variant of their input $x_t$, where $x_t$ is a noisy version of the input $x$. The corresponding
objective can be simplified to
\begin{eqnarray}
L_{DM}=\mathbb{E}_{x,\epsilon \sim \mathcal{N}\left( 0,1 \right) ,t}\left[ \| \epsilon - \epsilon_{\theta} \left( x_{t},t \right) \|_{2}^{2} \right]
\label{eqn:dm}
\end{eqnarray}
Diffusion  models  can learn the diverse data distribution in multiple domains. We aim to leverage this advantage and design more suitable generative conditions to address the issues of time delays between agents in collaborative perception.

\textbf{The Design of Condition. }
Inspired by the literature \cite{latent_diffusion}, diffusion models can leverage conditional mechanisms $p\left( x|y \right)$ to control the target generation process, thus improving the accuracy of the model. 
We find that in multi-agent collaborative perception, the information from individual agents can be  naturally used as conditional input to generate the final fused features. Thus, the generation of multi-agent features can be controlled by manipulating the input conditions $y$.

In practice, we select the information on the characteristics of the individual agents $N-1$ as condition $y_{cond}$. These informations are obtained by generating features $F_{j}$ through point cloud encoders, which are then transmitted to the Ego agent $i$, denoted as $\mathcal{M}_{j\rightarrow i}$. 
We strictly follow the standard DM model to add noise, while the difference is that we employ condition-modulated denoising, which is shown in Figure \ref{fig:all}. By progressively denoising the samples, we expect the conditional diffusion model to serve as a more flexible conditional generator, effectively suppressing time-delay-induced noise and thereby facilitating the learning of fine-grained object representations, such as precise boundaries and highly detailed shapes, particularly under noisy conditions.

Given the noisy feature $z_t$ and the condition $y_{cond}=\{ \mathcal{M}_{j\rightarrow i}\}$ at time step $t$, $z_t$ is further encoded and interacts with $y_{cond}$ through concatenation. This form $p\left( z_t |y_{cond} \right)$ can
be implemented with a conditional denoising autoencoder $\epsilon_{\theta} \left( z_{t},t,y_{cond} \right)$. 
A Unet-style structure \cite{unet}, whose components include an encoder and a decoder, severs as the denoising network.
Based on feature-conditioning pairs, we then learn the conditional LDM via
\begin{eqnarray}
L_{LDM}:=\mathbb{E}_{\mathcal E\left( x \right),y,\epsilon \sim \mathcal{N}\left( 0,1 \right) ,t}\left[ \parallel \epsilon -\epsilon_{\theta} \left( z_{t},t,y_{cond} \right) \parallel_{2}^{2} \right]
\label{eqn:diffusion_loss}
\end{eqnarray}
The definition of a symbols can be referred to the equation (\ref{eqn:dm}).
After generating the  feature map $ F^{\prime }_{i}$,
we decode them into the detection layer $\Phi_{Dec}\left( \cdot \right)  $ to obtain the final object detections $\mathcal{O}_{i}$.

\subsection{Training details and loss function}

Following common practice \cite{latent_diffusion, condition1}, the training process is divided into two stages. First, we train a perception compression model to construct a low-dimensional representation space. Second, a conditional diffusion model is trained within this space. For optimization, we adopt cross-entropy loss for classification ($\mathcal{L}_{cls}$) and weighted smooth L1 loss for regression ($\mathcal{L}_{reg}$). The total loss is the weighted sum of the diffusion loss in Equation (\ref{eqn:diffusion_loss})  and the detection loss.
 \begin{eqnarray}
\mathcal{L}_{total}=\mathcal{L}_{LDM}+\lambda_{cls} \mathcal{L}_{cls}+\lambda_{reg} \mathcal{L}_{reg}
\end{eqnarray}

\begin{table*}[]
 \caption{3D object detection performance on  DAIR-V2X\cite{dair}, V2XSet\cite{v2x-vit} and OPV2V \cite{opv2v} datasets. 
Experiments show that CoDiff achieves the overall best performance under various noise levels. The symbol '-' means the results are unavailable. }
 \setlength{\tabcolsep}{2.5pt}
 \begin{tabular}{llllllllllllllll}
\hline
\multicolumn{1}{l|}{Dataset}       & \multicolumn{5}{c}{DAIR-V2X}                                                                                                                                                                                          & \multicolumn{5}{c}{V2XSet}                                                                                                                                                                                            & \multicolumn{5}{c}{OPV2V}                                                                                                                                                                        \\ \hline
\multicolumn{1}{l|}{Method/Metric} & \multicolumn{15}{c}{AP@0.5 ↑}                                                                                                                                                                                                                                                                                                                                                                                                                                                                                                                                                                                                                      \\ \hline
\multicolumn{1}{l|}{Noise Level ($\sigma_{t} /\sigma_{r} \left( m/^{\circ }\right) $)}   & 0.0/0.0                              & 0.1/0.1                              & 0.2/0.2                              & 0.3/0.3                              & \multicolumn{1}{l|}{0.4/0.4}                              & 0.0/0.0                              & 0.1/0.1                              & 0.2/0.2                              & 0.3/0.3                              & \multicolumn{1}{l|}{0.4/0.4}                              & 0.0/0.0                              & 0.1/0.1                              & 0.2/0.2                              & 0.3/0.3                              & 0.4/0.4                              \\ \hline
\multicolumn{1}{l|}{F-Cooper \cite{f-cooper}}      & 73.4                                 & 70.2                                 & 72.3                                 & 69.5                                 & \multicolumn{1}{l|}{70.5}                                 & 78.3                                 & 77.7                                 & 76.3                                 & 73.5                                 & \multicolumn{1}{l|}{71.2}                                 & 83.4                                 & 82.6                                 & 78.8                                 & 72.3                                 & 68.1                                 \\
\multicolumn{1}{l|}{V2VNet \cite{v2vnet}}        & 66.0                                 & 65.7                                 & 65.5                                 & 64.9                                 & \multicolumn{1}{l|}{64.6}                                 & 87.1                                 & 86.6                                 & 86.0                                 & 84.3                                 & \multicolumn{1}{l|}{83.2}                                 & 94.2                                 & 93.9                                 & 93.8                                 & 93.0                                 & 92.9                                 \\
\multicolumn{1}{l|}{Self-ATT \cite{opv2v}}      & 70.5                                 & 70.4                                 & 70.3                                 & 69.8                                 & \multicolumn{1}{l|}{69.5}                                 & 87.6                                 & 87.3                                 & 86.8                                 & 85.9                                 & \multicolumn{1}{l|}{85.4}                                 & 94.3                                 & 93.9                                 & 93.3                                 & 92.1                                 & 91.5                                 \\
\multicolumn{1}{l|}{V2X-ViT \cite{v2x-vit}}       & 70.4                                 & 70.3                                 & 70.0                                 & 68.5                                 & \multicolumn{1}{l|}{68.9}                                 & 91.0                                 & 90.7                                 & 90.1                                 & 88.7                                 & \multicolumn{1}{l|}{86.9}                                 & 94.6                                 & 93.9                                 & 94.2                                 & 93.3                                 & 93.1                                 \\
\multicolumn{1}{l|}{CoBEVFlow \cite{CoBEVFlow}}     & 73.8                                 & -                                    & 73.2                                 & -                                    & \multicolumn{1}{l|}{70.3}                                 & -                                    & -                                    & -                                    & -                                    & \multicolumn{1}{l|}{-}                                    & -                                    & -                                    & -                                    & -                                    & -                                    \\
\multicolumn{1}{l|}{CoAlign \cite{CoAlign}}       & 74.6                                 & 74.5                                 & 73.8                                 & 72.4                                 & \multicolumn{1}{l|}{72.0}                                 & 91.9                                 & 91.6                                 & 90.9                                 & 89.7                                 & \multicolumn{1}{l|}{88.1}                                 & 96.6                                 & 96.5                                 & 96.2                                 & 95.9                                 & 95.8                                 \\
\multicolumn{1}{l|}{RoCo \cite{roco}}          & 76.3                                 & 74.9                                 & 74.8                                 & 73.1                                 & \multicolumn{1}{l|}{73.3}                                 & 91.9                                 & 91.7                                 & 91.0                                 & 89.6                                 & \multicolumn{1}{l|}{90.0}                                 & 96.6                                 & 96.6                                 & 96.6                                 & 95.8                                 & 95.7                                 \\ \hline
\multicolumn{1}{l|}{Ours (CoDiff)} & {\color[HTML]{000000} \textbf{77.4}} & {\color[HTML]{000000} \textbf{77.0}} & {\color[HTML]{000000} \textbf{75.4}} & {\color[HTML]{000000} \textbf{73.6}} & \multicolumn{1}{l|}{{\color[HTML]{000000} \textbf{73.4}}} & {\color[HTML]{000000} \textbf{92.1}} & {\color[HTML]{000000} \textbf{92.0}} & {\color[HTML]{000000} \textbf{91.0}} & {\color[HTML]{000000} \textbf{90.0}} & \multicolumn{1}{l|}{{\color[HTML]{000000} \textbf{90.0}}} & {\color[HTML]{000000} \textbf{97.0}} & {\color[HTML]{000000} \textbf{97.0}} & {\color[HTML]{000000} \textbf{96.7}} & {\color[HTML]{000000} \textbf{96.5}} & {\color[HTML]{000000} \textbf{96.0}} \\ \hline
\multicolumn{11}{l}{}                                                                                                                                                                                                                                                                                                                                                                                                                                                              & \multicolumn{5}{l}{}                                                                                                                                                                             \\ \hline
\multicolumn{1}{l|}{Method/Metric} & \multicolumn{15}{c}{AP@0.7 ↑}                                                                                                                                                                                                                                                                                                                                                                                                                                                                                                                                                                                                                      \\ \hline
\multicolumn{1}{l|}{Noise Level ($\sigma_{t} /\sigma_{r} \left( m/^{\circ }\right) $)}   & 0.0/0.0                              & 0.1/0.1                              & 0.2/0.2                              & 0.3/0.3                              & \multicolumn{1}{l|}{0.4/0.4}                              & 0.0/0.0                              & 0.1/0.1                              & 0.2/0.2                              & 0.3/0.3                              & \multicolumn{1}{l|}{0.4/0.4}                              & 0.0/0.0                              & 0.1/0.1                              & 0.2/0.2                              & 0.3/0.3                              & 0.4/0.4                              \\ \hline
\multicolumn{1}{l|}{F-Cooper \cite{f-cooper}}      & 55.9                                 & 53.4                                 & 55.2                                 & 53.7                                 & \multicolumn{1}{l|}{54.2}                                 & 48.6                                 & 47.3                                 & 46.0                                 & 44.1                                 & \multicolumn{1}{l|}{43.4}                                 & 60.2                                 & 55.3                                 & 50.4                                 & 46.8                                 & 41.2                                 \\
\multicolumn{1}{l|}{V2VNet \cite{v2vnet}}        & 48.6                                 & 48.4                                 & 48.3                                 & 48.0                                 & \multicolumn{1}{l|}{47.8}                                 & 64.6                                 & 63.5                                 & 62.0                                 & 57.2                                 & \multicolumn{1}{l|}{56.2}                                 & 85.4                                 & 85.1                                 & 84.8                                 & 84.0                                 & 83.7                                 \\
\multicolumn{1}{l|}{Self-ATT \cite{opv2v}}      & 52.2                                 & 52.1                                 & 52.0                                 & 51.7                                 & \multicolumn{1}{l|}{51.7}                                 & 67.6                                 & 67.2                                 & 66.2                                 & 65.4                                 & \multicolumn{1}{l|}{65.1}                                 & 82.7                                 & 81.6                                 & 80.4                                 & 79.5                                 & 78.0                                 \\
\multicolumn{1}{l|}{V2X-ViT \cite{v2x-vit}}       & 53.1                                 & 53.2                                 & 52.9                                 & 52.9                                 & \multicolumn{1}{l|}{52.5}                                 & 80.3                                 & 79.1                                 & 76.8                                 & 74.2                                 & \multicolumn{1}{l|}{71.8}                                 & 85.6                                 & 85.0                                 & 85.1                                 & 84.3                                 & 84.1                                 \\
\multicolumn{1}{l|}{CoBEVFlow \cite{CoBEVFlow}}     & 59.9                                 & -                                    & 57.9                                 & -                                    & \multicolumn{1}{l|}{56.0}                                 & -                                    & -                                    & -                                    & -                                    & \multicolumn{1}{l|}{-}                                    & -                                    & -                                    & -                                    & -                                    & -                                    \\
\multicolumn{1}{l|}{CoAlign \cite{CoAlign}}       & 60.4                                 & 60.0                                 & 58.8                                 & 58.3                                 & \multicolumn{1}{l|}{57.9}                                 & 80.5                                 & 79.4                                 & 77.3                                 & 75.1                                 & \multicolumn{1}{l|}{73.0}                                 & 91.2                                 & 90.8                                 & 90.0                                 & 89.4                                 & 88.9                                 \\
\multicolumn{1}{l|}{RoCo \cite{roco}}          & 62.0                                 & 59.8                                 & 59.4                                 & 58.4                                 & \multicolumn{1}{l|}{{\color[HTML]{000000} \textbf{58.4}}} & 80.5                                 & 79.4                                 & 77.4                                 & 76.1                                 & \multicolumn{1}{l|}{{\color[HTML]{000000} \textbf{77.3}}} & 91.3                                 & 90.9                                 & 90.1                                 & 89.4                                 & 89.1                                 \\ \hline
\multicolumn{1}{l|}{Ours (CoDiff)} & {\color[HTML]{000000} \textbf{62.7}} & {\color[HTML]{000000} \textbf{61.6}} & {\color[HTML]{000000} \textbf{59.6}} & {\color[HTML]{000000} \textbf{58.4}} & \multicolumn{1}{l|}{{\color[HTML]{000000} \underline{58.3}}}          & {\color[HTML]{000000} \textbf{82.3}} & {\color[HTML]{000000} \textbf{80.7}} & {\color[HTML]{000000} \textbf{77.5}} & {\color[HTML]{000000} \textbf{77.2}} & \multicolumn{1}{l|}{{\color[HTML]{000000} \underline{76.3}}}          & {\color[HTML]{000000} \textbf{91.7}} & {\color[HTML]{000000} \textbf{91.3}} & {\color[HTML]{000000} \textbf{91.0}} & {\color[HTML]{000000} \textbf{90.0}} & {\color[HTML]{000000} \textbf{89.3}} \\ \hline
\end{tabular}
\label{tab-all}
\end{table*}

\begin{figure}[htbp]
\centering
\subfigure[V2X-VIT]{
\includegraphics[width=0.480\linewidth]{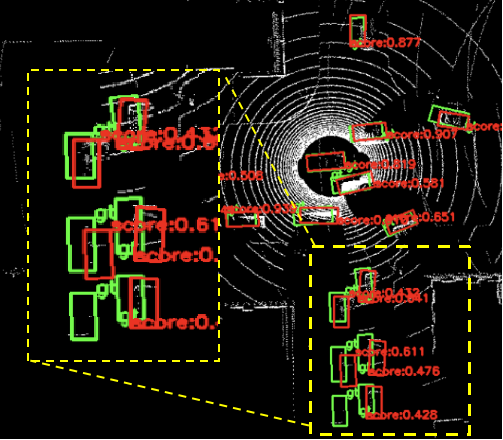}}
\subfigure[CoAlign]{
\includegraphics[width=0.46\linewidth]{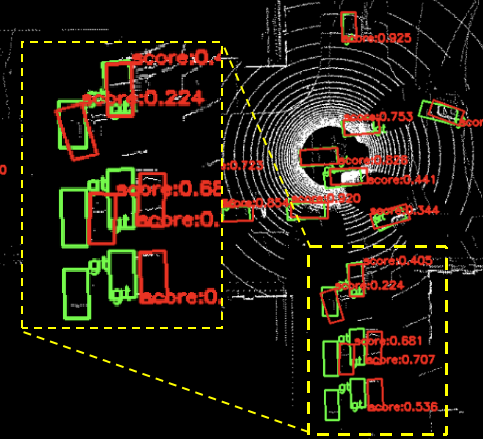}}
\vfill
\subfigure[RoCo]{
\includegraphics[width=0.48\linewidth]{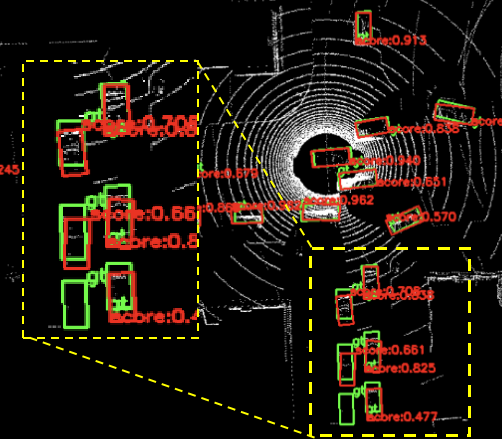}}
\subfigure[Ours]{
\includegraphics[width=0.46\linewidth]{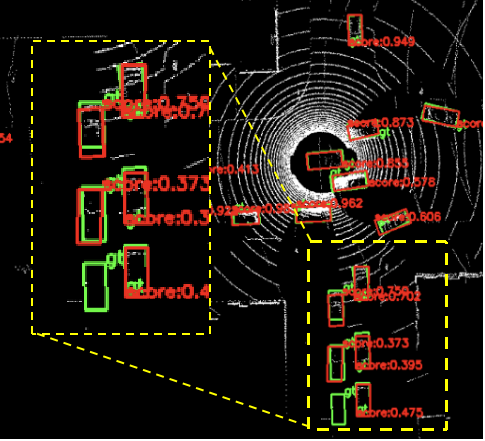}}
\caption{Visualization of detection results for V2X-ViT, CoAlign, RoCo and  our CoDiff with the noisy level $\sigma^2_{t} /\sigma^2_{r} \left( m/^{\circ }\right) $ of 0.4/0.4 and 100ms time delay  on V2XSet dateset.  CoDiff achieves much more precise detection.
}
\label{fig-view}
\end{figure}

\section{Experiment Result}
We validate our CoDiff  on both simulated and real-world scenarios. Following the literature, the detection performance are evaluated by using Average Precision (AP) at Intersection-over-Union (IoU) thresholds of 0.50 and 0.70.

\subsection{Datasets}

\textbf{DAIR-V2X \cite{dair}.} DAIR-V2X is a large-scale vehicle-infrastructure cooperative perception dataset containing over 100 scenes and 18,000 data pairs, featuring two agents: vehicle and road-side unit (RSU), capturing simultaneous data from infrastructure and vehicle sensors at an equipped intersection as an autonomous vehicle passes through. 
\textbf{V2XSet \cite{v2x-vit}.}  V2XSet is built with the co-simulation of OpenCDA \cite{opencda} and CARLA \cite{carla}.
It is a large-scale simulated dataset designed for Vehicle-to-Infrastructure (V2X) communication. The dataset consists of a total of 11,447 frames, and the train/validation/test splits are 6,694/1,920/2,833, respectively.  
\textbf{OPV2V \cite{opv2v}.} OPV2V is a large-scale dataset specifically designed for vehicle-to-vehicle (V2V) communication. It is jointly developed using the CARLA and OpenCDA simulation tools\cite{carla}. It includes 12K frames of 3D point clouds and RGB images with 230K annotated 3D boxes.

\begin{figure}[t]
\centering
\includegraphics[width=1\columnwidth]{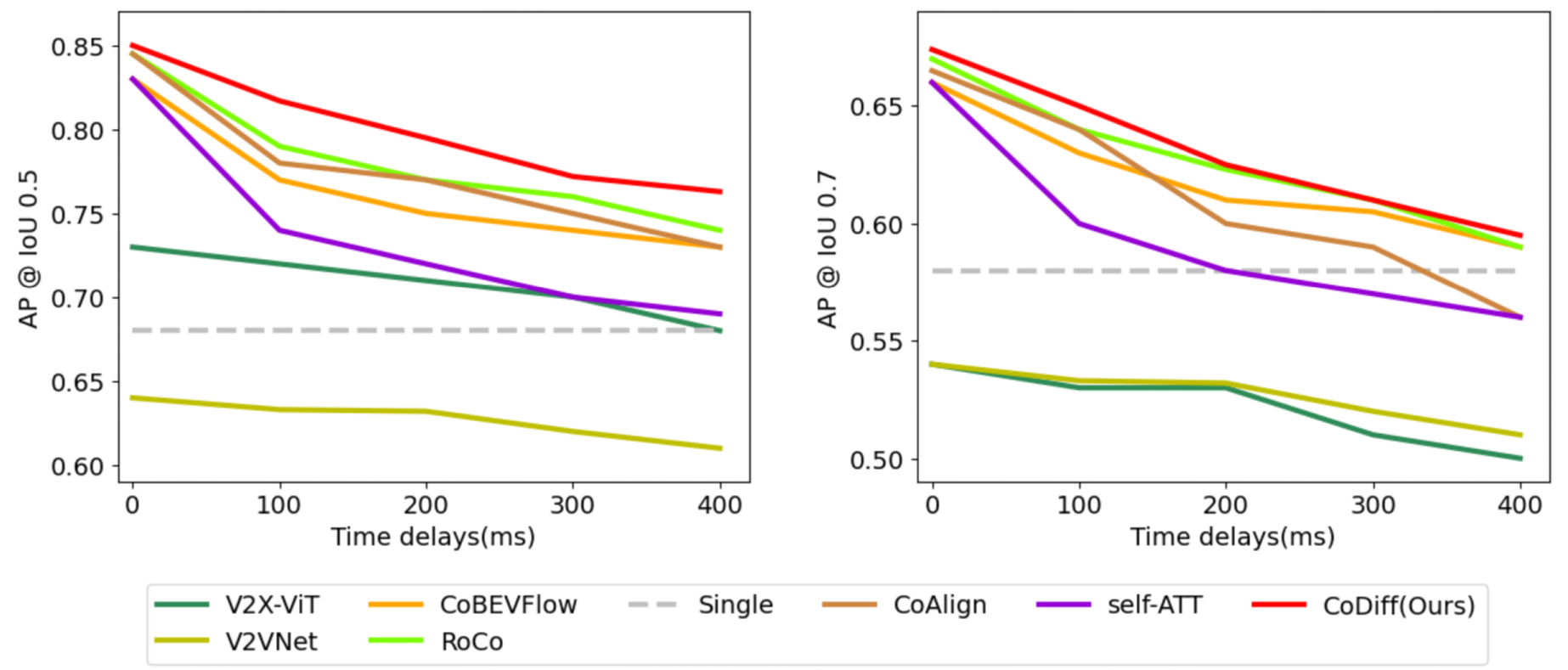} 
\caption{Comparison of the performance of CoDiff and other baseline methods on DAIR-V2X dataset under the time delay from 0 to 400ms.}
\label{fig-time}
\end{figure}

\subsection{Implementation Details}

We keep PointPillar \cite{pointpillars} as LiDAR encoder backbone. Converts the
LiDAR point cloud into voxels with a resolution of 0.4 m for both height and width.
In bipartite graph matching, we set the similarity threshold $\tau_1=0.5$ for Eq.(\ref{eq:all}) and $\lambda =1$ for Eq.(\ref{eqn:Simila}). 
In pose graph optimization, the Levenberg-Marquardt algorithm\cite{levenberg} is used to solve the least squares optimization problem, with the maximum number of iterations set to 1000.
We train an auto-encoder to compress the features to an 8-dimensional space with a 32x compression rate. The diffusion model is trained using the Adam optimizer with a learning rate of 0.0002 and weight decay of 0.001. The training is conducted over 20 epochs with a total of 500 steps, adding diffusion conditions at each step. Finally, the object detection model is fine-tuned for 10 epochs.
To simulate pose inconsistencies and time delays, we follow the noisy settings  in RoCo \cite{roco} and the delay setting in CoBEVFlow \cite{v2x-vit} during the training process.  We add Gaussian noise $N\left( 0,\sigma^2_{t} \right)   $ on $x,y$ and $N\left( 0,\sigma^2_{r} \right)  $ on $\theta $, where $x,y,\theta $ are the coordinates of the 2D centers of a vechicle and the yaw angle of accurate global poses. The time delay is set to 100 ms. 
All models are trained on six NVIDIA RTX 2080Ti GPUs. 

\subsection{Quantitative evaluation}

To validate the overall performance of CoDiff in 3D object detection, 
we compare it with seven state-of-the-art methods on the three datasets:  F-Cooper \cite{f-cooper}, V2VNet \cite{v2vnet}, Self-ATT \cite{opv2v}, V2X-ViT \cite{v2x-vit}, CoBEVFlow \cite{CoBEVFlow} , CoAlign \cite{CoAlign} and RoCo \cite{roco}. 
For a fair comparison, all models take the same 3D point clouds as input data. All methods use the same feature encoder based on PointPillars \cite{pointpillars}. 
For a fair comparison, all models take the same 3D point clouds as input data. All methods use the same feature encoder based on PointPillars \cite{pointpillars}. 
Table \ref{tab-all} shows  the AP at IoU threshold of 0.5 and 0.7 in DAIR-V2X, V2XSet and OPV2V dataset. We see that CoDiff significantly outperforms the previous methods at various noise levels across the three datasets and the leading gap is larger when the noise level is higher. 

We further investigate the impact of the time delay in the range [0, 400] ms.
Fig. \ref{fig-time} shows the detection performances  of the proposed CoDiff and the baseline methods under varying levels of time delay on  DAIR-V2X.
We see that  the proposed CoDiff achieves the best performance at all time delay settings, as shown by the red line in the graph. In the case of a time delay of 200ms, our approach achieves an improvement of 2. 6\% (82. 0\% $vs.$ 79. 4\%) and 0. 5\% (59. 5\% $vs.$ 59. 0\%) over RoCo for AP@0.5/0.7. As the delay increases, all methods experience varying degrees of performance degradation, however, even under a 400 ms delay, our approach still achieves superior detection accuracy.


\begin{figure}[t]
\centering
\includegraphics[width=1\columnwidth]{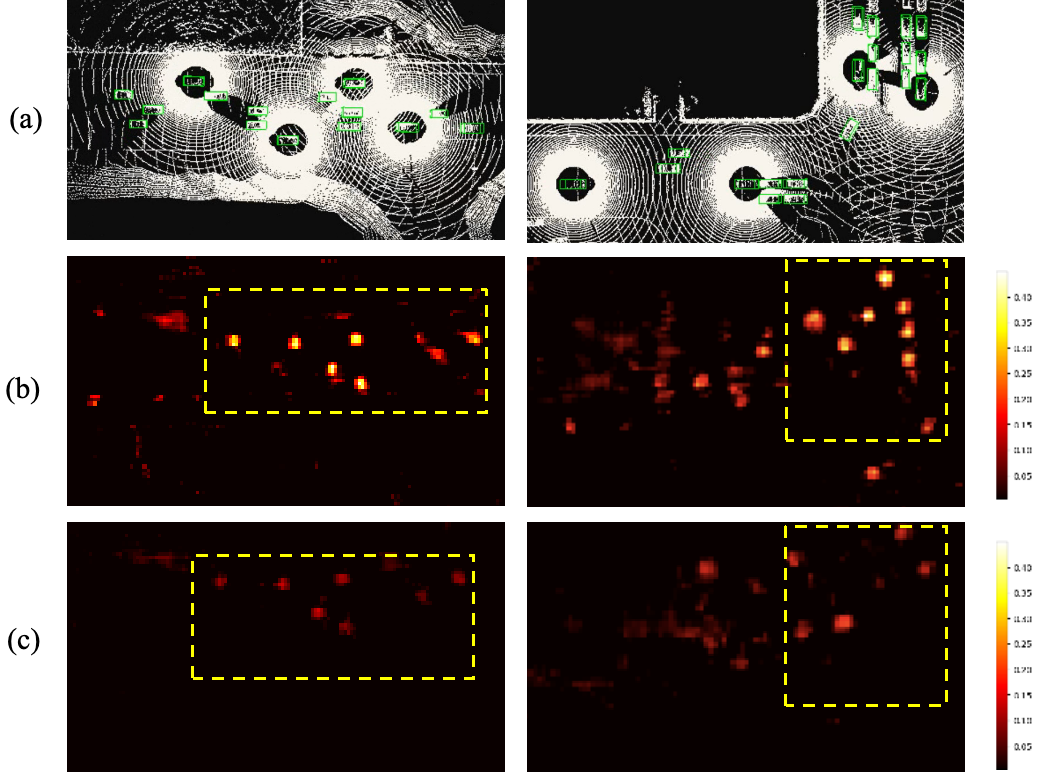} 
\caption{Visualization of attention maps with diffusion model (b) and without the  diffusion model (c). (a) is the ground truth of the different road types.}
\label{fig-att}
\end{figure}

\begin{table}[]
\centering
\caption{Component ablation study.}
\begin{tabular}{cc|cc|cc}
\hline
\multicolumn{2}{c|}{Modules} & \multicolumn{2}{c|}{AP@0.5}                                                 & \multicolumn{2}{c}{AP@0.7}                                                  \\ \hline
PCM     & TCM     & 0.2/0.2                              & 0.4/0.4                              & 0.2/0.2                              & 0.4/0.4                              \\ \hline
\textbf{×}            & \textbf{×}             & 73.8                                 & 72.0                                 & 58.8                                 & 57.9                                 \\
\checkmark            & \textbf{×}             & 74.8                                 & 73.3                                 & 59.4                                 & 58.4                                 \\
\textbf{×}            & \checkmark             & 75.0                                 & 72.1                                 & 59.8                                 & 57.4                                 \\
\checkmark            & \checkmark             & {\color[HTML]{000000} \textbf{75.4}} & {\color[HTML]{000000} \textbf{73.6}} & {\color[HTML]{000000} \textbf{60.0}} & {\color[HTML]{000000} \textbf{59.0}} \\ \hline
\end{tabular}
\label{tab4}
\end{table}

\subsection{Qualitative evaluation}
\textbf{Detection visualization.}
We  show the 3D detection results in the Bird's-eye-view (BEV) format on the V2XSet. 
The degree of overlapping of these boxes reflects the performance of a method.  
Figure \ref{fig-view}  depicts the detection results of V2X-ViT, CoAlign, RoCo and the proposed CoDiff at an intersection to validate the effectiveness of our method. 
We set the noise level and time delay of $0.4m/0.4^{\circ }$ and 100ms to produce high noises to make the perception task challenging.  From the figure, it can be observed that V2X-ViT has many missed detections, while CoAlign \cite{CoAlign} and RoCo \cite{roco} generate many predictions with relatively large offsets. In contrast, our CoDiff demonstrates strong performance in the presence of  pose inconsistencies and time delays.

\textbf{Attention map visualization.}
To better understand the advantages of the diffusion model, we visualize the multi-agent fused feature maps learned by different methods in Fig. \ref{fig-att}. In these visualizations, brighter areas indicate higher feature intensity, which is more beneficial for subsequent detection. Fig. \ref{fig-att}(b) shows the feature map generated by the conditional DPM, while Fig. \ref{fig-att}(c) visualizes the feature map learned by the other multi-scale feature fusion method \cite{v2x-vit,where2comm}. It is evident from the figures that the feature intensity is higher with the conditional DPM, and the features at greater distances are also clearer (highlighted in the yellow boxes).

\begin{table}[t]
\centering
\caption{Selection of the compression rate $\tau$ in CoDiff.}
\begin{tabular}{c|c|l}
\hline
Compression rate $\tau$ & Channel $c$  & mAP                                  \\ \hline
8x                 & 32 & 82.5                                 \\
16x               & 16 & 83.5                                 \\
32x                & 8  & {\color[HTML]{000000} \textbf{84.6}} \\
64x                & 4  & 82.5                                 \\ \hline
\end{tabular}
 \label{fig-compression}
\end{table}

\begin{table}[]
\caption{Selection of the threshold value $\tau_2$ in Eq. (\ref{eqn:tau_2}).}
\setlength{\tabcolsep}{2pt}
\resizebox{\linewidth}{!}{
\begin{tabular}{c|cccccccc}
\hline
                                   & \multicolumn{4}{c}{DAIR-V2X}                               & \multicolumn{4}{c}{V2XSet}            \\ \hline
Threshold/Metric                   & \multicolumn{8}{c}{AP@0.7}                                                                         \\ \hline
Noise Level  & 0.2/0.2 & 0.4/0.4 & 0.6/0.6 & \multicolumn{1}{l|}{0.8/0.8} & 0.2/0.2 & 0.4/0.4 & 0.6/0.6 & 0.8/0.8 \\ \hline
$\tau_2$ = 1                              & 58.8    & 57.6    & 56.9    & \multicolumn{1}{l|}{56.4}    & 77.1    & 72.5    & 70.4    & 65.7    \\
$\tau_2$ = 2                              & 58.9    & 57.9    & 57.2    & \multicolumn{1}{l|}{56.9}    & {\color[HTML]{000000} \textbf{77.4}} & {\color[HTML]{000000} \textbf{77.3}} & {\color[HTML]{000000} \textbf{71.0}}    & 68.1    \\
$\tau_2$ = 3                              & {\color[HTML]{000000} \textbf{59.4}} & {\color[HTML]{000000} \textbf{58.4}} & {\color[HTML]{000000} \textbf{58.2}}    & \multicolumn{1}{l|}{{\color[HTML]{000000} \textbf{57.8}}}    & 77.1    & 73.1    & 70.2    & {\color[HTML]{000000} \textbf{68.9}}    \\
$\tau_2$ = 4                              & 57.9    & 57.6    & 57.6    & \multicolumn{1}{l|}{57.5}    & 75.9    & 72.0    & 69.7    & 68.3    \\ \hline
\end{tabular}
}
\label{tab2}
\end{table}


\subsection{Ablation Studies}
\textbf{Contribution of major components in CoDiff.} 
Now we investigate the effectiveness of individual components in CoDiff. We evaluate the impact of each component by progressively adding PCM and TCM. As Table~\ref{tab4} demonstrates, all the modules are beneficial to the performance gains, among which the pose adjustment module is particularly helpful in enhancing detection accuracy under high-noise conditions.

\textbf{Selection of the optimal compression rate $\tau$.}
The selection of the optimal compression rate in CoDiff is crucial, as it directly impacts the quality of information shared between agents and the overall detection performance.
To determine the optimal compression rate $\tau$ during the perceptual compression, we conduct ablation experiments on V2XSet dataset, 
as shown in Table \ref{fig-compression}.  
We found that setting $\tau$ to 32 and compressing the feature dimensions  to 8 achieved the optimal mAP.

\begin{table}[htbp]
\centering
\caption{Contribution of graph similarity.}
 \setlength{\tabcolsep}{3.5pt}
\begin{tabular}{ccc|cc|cc}
    \hline
    \multicolumn{3}{c|}{Modules}                    & \multicolumn{2}{c|}{AP@0.5} & \multicolumn{2}{c}{AP@0.7}  \\ \hline
    \multicolumn{1}{c|}{Matching} & distance & edge & 0.2/0.2 & 0.4/0.4 & 0.2/0.2 & 0.4/0.4 \\ \hline
    \multicolumn{1}{c|}{\textbf{×}}        & \textbf{×}        & \textbf{×}    & 73.8    & 72.0    & 58.8    & 57.9    \\
    \multicolumn{1}{c|}{\checkmark }        & \checkmark         & \textbf{×}     & 73.0    & 71.0    & 58.8    & 58.0    \\
    \multicolumn{1}{c|}{\checkmark }        & \textbf{×}         & \checkmark     & 73.3    & 72.1    & 59.0    & 58.1    \\
    \multicolumn{1}{c|}{\checkmark }        & \checkmark         & \checkmark     & {\color[HTML]{000000} \textbf{74.8}}    & {\color[HTML]{000000} \textbf{73.3}}    & {\color[HTML]{000000} \textbf{59.4}}    & {\color[HTML]{000000} \textbf{58.4}}    \\ \hline
\end{tabular}
\label{tab3}
\end{table}

\begin{table}[htbp]
\caption{Ablation study on V2XSet val set.}
 \setlength{\tabcolsep}{2pt}
\resizebox{\linewidth}{!}{
\begin{tabular}{ccc|ccc|lll}
\hline
\multicolumn{3}{c|}{Modules}                                       & \multicolumn{3}{c|}{AP@0.5}                                                                                        & \multicolumn{3}{c}{AP@0.7}                                                                                         \\ \hline
\multicolumn{1}{l|}{Condition} & \multicolumn{1}{l|}{Add.} & Concat. & 0.0/0.0                              & 0.2/0.2                              & 0.4/0.4                              & 0.0/0.0                              & 0.2/0.2                              & 0.4/0.4                              \\ \hline
\multicolumn{1}{l|}{\textbf{×}}         & \multicolumn{1}{l|}{\textbf{×}}   & \textbf{×}      & 94.3                                 & 92.1                                 & 89.5                                 & 86.5                                 & 83.0                                 & 78.6                                 \\
\multicolumn{1}{l|}{\checkmark}         & \multicolumn{1}{l|}{\checkmark}   & \textbf{×}      & 95.2                                 & 92.6                                 & 90.8                                 & 87.4                                 & 84.1                                 & 79.5                                 \\
\multicolumn{1}{l|}{\checkmark}         & \multicolumn{1}{l|}{\textbf{×}}   & \checkmark      & {\color[HTML]{000000} \textbf{96.2}} & {\color[HTML]{000000} \textbf{95.4}} & {\color[HTML]{000000} \textbf{91.7}} & {\color[HTML]{000000} \textbf{89.7}} & {\color[HTML]{000000} \textbf{86.0}} & {\color[HTML]{000000} \textbf{80.4}} \\ \hline
\end{tabular}
}
 \label{fig-condition}
\end{table}

\textbf{Selection of the threshold value $\tau_2$.}
To identify the optimal value for the threshold $\tau_2$ in Eq.~(\ref{eqn:tau_2}) during graph initialization, we perform ablation studies on multiple datasets, as shown in Table~\ref{tab2}. On the DAIR-V2X dataset, we observe that setting $\tau_2$ to 3 yields the best detection accuracy. When $\tau_2$ is increased beyond this point, performance degrades—primarily because vehicles in real-world environments tend to keep safe distances, and an overly small threshold would hinder effective initial matching.
We also conduct the same analysis on a simulated dataset. The results show that when the noise level is below $0.8\,\text{m}/0.8^\circ$, the best detection performance occurs at $\tau_2 = 2\,\text{m}$. However, as the noise level rises to $0.8\,\text{m}/0.8^\circ$, the optimal threshold shifts to $\tau_2 = 3$. This change is attributed to the characteristics of the V2XSet dataset, where vehicles are more numerous and densely packed, resulting in generally shorter inter-vehicle distances compared to real-world data.

\textbf{Contribution of graph similarity.} 
In Table~\ref{tab3}, we apply the similarity metrics defined in Eq.~(\ref{eqn:edge}) and Eq.~(\ref{eqn:distance}) to the object matching process on the DAIR-V2X dataset. To assess the individual contribution of each similarity measure to the overall graph matching performance, we progressively introduce (i) edge-based similarity and (ii) distance-based similarity. The first row in Table~\ref{tab3} serves as the baseline, where no similarity-based matching is applied. Experimental findings demonstrate that incorporating either form of similarity leads to notable improvements in detection accuracy.

\begin{table}[htbp]
\centering
\caption{Performance comparison of different
sampling methods at varying steps in the proposed CoDiff.}
 \setlength{\tabcolsep}{3.5pt}
\begin{tabular}{l|ccccc}
\hline
Method/Metric & \multicolumn{5}{c}{AP@0.7}       \\ \hline
Sampling Step & 1    & 2    & 4    & 8    & 10   \\ \hline
DDPM  \cite{DDPM}        & 84.7 & 85.3 & 85.7 & {\color[HTML]{000000} \textbf{86.0}} & 85.4 \\
DDIM  \cite{DDIM}        &  85.8    &   85.4   &  85.6    &  85.8    &   85.3   \\ \hline
\end{tabular}
 \label{tab-sample}
\end{table}

\begin{table}[htbp]
\centering
\caption{Inference time and Parameters.}
\begin{tabular}{l|ll}
\hline
Method  & Interfence\_time & Parameters \\ \hline
CoDiff (DDPM)   & 127.7 ms          & 769.17  MB \\
CoDiff (DDIM) & 125.8  ms          & 776.91  MB \\ \hline
\end{tabular}
 \label{tab-time}
\end{table}

\textbf{Effectiveness of the conditional diffusion module.} 
Table \ref{fig-condition} shows the results of using or not using conditions, as well as how the conditions were applied in the model. Please note that: (i) The conditions enhance the stability of our porposed method; (ii) Using Concat for condition fusion is superior to directly applying element-wise addition.
In testing, we used two samplers: DDPM \cite{DDPM} and DDIM \cite{DDIM}. To achieve optimal performance, we tested the effect of different samplers and sampling steps on the results. The experimental results are shown in Table \ref{tab-sample}. We selected DDPM with 8 sampling steps as the default solver in the experiments due to its simplicity and efficiency.
We  also measure the inference time and the parameters of our proposed CoDiff on a single
NVIDIA 2080Ti GPU in Table \ref{tab-time}. At 1-step DDPM sampling, CoDiff can reach the inference time  of 127.7 (ms) with 769.17 (MB) parameters. 
At 1-step DDIM sampling, CoDiff can reach the inference time  of 125.8 (ms) with 776.91 (MB) parameters.

\section{CONCLUSION}
This paper proposes a novel robust collaborative perception framework for 3D object detection, called CoDiff, which leverages conditional diffusion models to address the problem of feature misalignment. CoDiff primarily consists of two modules: PCM and TCM. The PCM establishes reliable correspondences between objects and performs pose calibration through graph optimization, while the TCM employs a conditional diffusion model to remove delay-induced noise, thereby generating highly refined multi-agent fused features. CoDiff does not require any ground truth pose and time supervision, making it highly practical. 
\color{black}
Comprehensive experiments demonstrate that our method achieves outstanding performance across all settings and exhibits exceptional robustness under extreme noise conditions. 
In future work, we aim to further explore the potential of CoDiff and expand its application to a broader range of perception tasks.

\section*{Acknowledgments}
This work was supported by the National Natural Science Foundation of China Grant No. 61972404, 12071478, the Public Computing Cloud of Renmin University of China, and the Blockchain Lab of the School of Information, Renmin University of China.

\bibliographystyle{IEEEtran}
\bibliography{sample-base}

\begin{thebibliography}{10}
\providecommand{\url}[1]{#1}
\csname url@samestyle\endcsname
\providecommand{\newblock}{\relax}
\providecommand{\bibinfo}[2]{#2}
\providecommand{\BIBentrySTDinterwordspacing}{\spaceskip=0pt\relax}
\providecommand{\BIBentryALTinterwordstretchfactor}{4}
\providecommand{\BIBentryALTinterwordspacing}{\spaceskip=\fontdimen2\font plus
\BIBentryALTinterwordstretchfactor\fontdimen3\font minus \fontdimen4\font\relax}
\providecommand{\BIBforeignlanguage}[2]{{%
\expandafter\ifx\csname l@#1\endcsname\relax
\typeout{** WARNING: IEEEtran.bst: No hyphenation pattern has been}%
\typeout{** loaded for the language `#1'. Using the pattern for}%
\typeout{** the default language instead.}%
\else
\language=\csname l@#1\endcsname
\fi
#2}}
\providecommand{\BIBdecl}{\relax}
\BIBdecl

\bibitem{pointpillars}
A.~H. Lang, S.~Vora, H.~Caesar, L.~Zhou, J.~Yang, and O.~Beijbom, ``Pointpillars: Fast encoders for object detection from point clouds,'' in \emph{Proceedings of the IEEE/CVF conference on computer vision and pattern recognition}, 2019, pp. 12\,697--12\,705.

\bibitem{3d}
S.~Chen, B.~Liu, C.~Feng, C.~Vallespi-Gonzalez, and C.~Wellington, ``3d point cloud processing and learning for autonomous driving: Impacting map creation, localization, and perception,'' \emph{IEEE Signal Processing Magazine}, vol.~38, no.~1, pp. 68--86, 2020.

\bibitem{pointnet++}
C.~R. Qi, L.~Yi, H.~Su, and L.~J. Guibas, ``Pointnet++: Deep hierarchical feature learning on point sets in a metric space,'' \emph{Advances in neural information processing systems}, vol.~30, 2017.

\bibitem{votenet++}
Z.~Ding and M.~Niethammer, ``Votenet++: Registration refinement for multi-atlas segmentation,'' in \emph{2021 IEEE 18th International Symposium on Biomedical Imaging (ISBI)}.\hskip 1em plus 0.5em minus 0.4em\relax IEEE, 2021, pp. 275--279.

\bibitem{TMM1}
M.~Dang, G.~Liu, H.~Li, D.~Wang, R.~Pan, and Q.~Wang, ``Pra-det: Anchor-free oriented object detection with polar radius representation,'' \emph{IEEE Transactions on Multimedia}, vol.~27, pp. 145--157, 2025.

\bibitem{TMM2}
S.~Zhang, D.~Kong, Y.~Xing, Y.~Lu, L.~Ran, G.~Liang, H.~Wang, and Y.~Zhang, ``Frequency-guided spatial adaptation for camouflaged object detection,'' \emph{IEEE Transactions on Multimedia}, vol.~27, pp. 72--83, 2025.

\bibitem{I1}
Y.~Hu, S.~Fang, W.~Xie, and S.~Chen, ``Aerial monocular 3d object detection,'' \emph{IEEE Robotics and Automation Letters}, vol.~8, no.~4, pp. 1959--1966, 2023.

\bibitem{I2}
S.~Chen, B.~Liu, C.~Feng, C.~Vallespi-Gonzalez, and C.~Wellington, ``3d point cloud processing and learning for autonomous driving: Impacting map creation, localization, and perception,'' \emph{IEEE Signal Processing Magazine}, vol.~38, no.~1, pp. 68--86, 2020.

\bibitem{D1}
J.~Ho, A.~Jain, and P.~Abbeel, ``Denoising diffusion probabilistic models,'' \emph{Advances in neural information processing systems}, vol.~33, pp. 6840--6851, 2020.

\bibitem{D2}
J.~Sohl-Dickstein, E.~Weiss, N.~Maheswaranathan, and S.~Ganguli, ``Deep unsupervised learning using nonequilibrium thermodynamics,'' in \emph{International conference on machine learning}.\hskip 1em plus 0.5em minus 0.4em\relax PMLR, 2015, pp. 2256--2265.

\bibitem{TMM3}
Y.~Mei, J.~Sun, Z.~Peng, F.~Deng, G.~Wang, and J.~Chen, ``Rog-sam: A language-driven framework for instance-level robotic grasping detection,'' \emph{IEEE Transactions on Multimedia}, vol.~27, pp. 3057--3068, 2025.

\bibitem{TMM4}
Z.~Liu, Y.~Shang, T.~Li, G.~Chen, Y.~Wang, Q.~Hu, and P.~Zhu, ``Robust multi-drone multi-target tracking to resolve target occlusion: A benchmark,'' \emph{IEEE Transactions on Multimedia}, vol.~25, pp. 1462--1476, 2023.

\bibitem{qi2017pointnet}
C.~R. Qi, H.~Su, K.~Mo, and L.~J. Guibas, ``Pointnet: Deep learning on point sets for 3d classification and segmentation,'' in \emph{Proceedings of the IEEE conference on computer vision and pattern recognition}, 2017, pp. 652--660.

\bibitem{huang}
Z.~Huang, Y.~Wang, J.~Wen, P.~Wang, and X.~Cai, ``An object detection algorithm combining semantic and geometric information of the 3d point cloud,'' \emph{Advanced Engineering Informatics}, vol.~56, p. 101971, 2023.

\bibitem{huang2}
Z.~Huang, Y.~Wang, X.~Tang, and H.~Sun, ``Boundary-aware set abstraction for 3d object detection,'' in \emph{2023 International Joint Conference on Neural Networks (IJCNN)}.\hskip 1em plus 0.5em minus 0.4em\relax IEEE, 2023, pp. 01--07.

\bibitem{TMM5}
T.~Xie, L.~Wang, K.~Wang, R.~Li, X.~Zhang, H.~Zhang, L.~Yang, H.~Liu, and J.~Li, ``Farp-net: Local-global feature aggregation and relation-aware proposals for 3d object detection,'' \emph{IEEE Transactions on Multimedia}, vol.~26, pp. 1027--1040, 2024.

\bibitem{HEAL}
Y.~Lu, Y.~Hu, Y.~Zhong, D.~Wang, S.~Chen, and Y.~Wang, ``An extensible framework for open heterogeneous collaborative perception,'' \emph{arXiv preprint arXiv:2401.13964}, 2024.

\bibitem{where2comm}
Y.~Hu, S.~Fang, Z.~Lei, Y.~Zhong, and S.~Chen, ``Where2comm: Communication-efficient collaborative perception via spatial confidence maps,'' \emph{Advances in neural information processing systems}, vol.~35, pp. 4874--4886, 2022.

\bibitem{roco}
Z.~Huang, S.~Wang, Y.~Wang, W.~Li, D.~Li, and L.~Wang, ``Roco: Robust cooperative perception by iterative object matching and pose adjustment,'' in \emph{ACM Multimedia 2024}.

\bibitem{SycNet}
Z.~Lei, S.~Ren, Y.~Hu, W.~Zhang, and S.~Chen, ``Latency-aware collaborative perception,'' in \emph{European Conference on Computer Vision}.\hskip 1em plus 0.5em minus 0.4em\relax Springer, 2022, pp. 316--332.

\bibitem{CoBEVFlow}
S.~Wei, Y.~Wei, Y.~Hu, Y.~Lu, Y.~Zhong, S.~Chen, and Y.~Zhang, ``Asynchrony-robust collaborative perception via bird's eye view flow,'' \emph{arXiv e-prints}, pp. arXiv--2309, 2023.

\bibitem{adversarial}
J.~Tu, T.~Wang, J.~Wang, S.~Manivasagam, M.~Ren, and R.~Urtasun, ``Adversarial attacks on multi-agent communication,'' in \emph{Proceedings of the IEEE/CVF International Conference on Computer Vision}, 2021, pp. 7768--7777.

\bibitem{CoAlign}
Y.~Lu, Q.~Li, B.~Liu, M.~Dianati, C.~Feng, S.~Chen, and Y.~Wang, ``Robust collaborative 3d object detection in presence of pose errors,'' in \emph{2023 IEEE International Conference on Robotics and Automation (ICRA)}.\hskip 1em plus 0.5em minus 0.4em\relax IEEE, 2023, pp. 4812--4818.

\bibitem{v2x-vit}
R.~Xu, H.~Xiang, Z.~Tu, X.~Xia, M.-H. Yang, and J.~Ma, ``V2x-vit: Vehicle-to-everything cooperative perception with vision transformer,'' in \emph{European conference on computer vision}.\hskip 1em plus 0.5em minus 0.4em\relax Springer, 2022, pp. 107--124.

\bibitem{opv2v}
R.~Xu, H.~Xiang, X.~Xia, X.~Han, J.~Li, and J.~Ma, ``Opv2v: An open benchmark dataset and fusion pipeline for perception with vehicle-to-vehicle communication,'' in \emph{2022 International Conference on Robotics and Automation (ICRA)}.\hskip 1em plus 0.5em minus 0.4em\relax IEEE, 2022, pp. 2583--2589.

\bibitem{dair}
H.~Yu, Y.~Luo, M.~Shu, Y.~Huo, Z.~Yang, Y.~Shi, Z.~Guo, H.~Li, X.~Hu, J.~Yuan \emph{et~al.}, ``Dair-v2x: A large-scale dataset for vehicle-infrastructure cooperative 3d object detection,'' in \emph{Proceedings of the IEEE/CVF Conference on Computer Vision and Pattern Recognition}, 2022, pp. 21\,361--21\,370.

\bibitem{syncnet}
A.~Raina and V.~Arora, ``Syncnet: Using causal convolutions and correlating objective for time delay estimation in audio signals,'' \emph{arXiv preprint arXiv:2203.14639}, 2022.

\bibitem{cobevt}
R.~Xu, Z.~Tu, H.~Xiang, W.~Shao, B.~Zhou, and J.~Ma, ``Cobevt: Cooperative bird's eye view semantic segmentation with sparse transformers,'' \emph{arXiv preprint arXiv:2207.02202}, 2022.

\bibitem{D3}
Y.~Song and S.~Ermon, ``Generative modeling by estimating gradients of the data distribution,'' \emph{Advances in neural information processing systems}, vol.~32, 2019.

\bibitem{D4}
Y.~Song, J.~Sohl-Dickstein, D.~P. Kingma, A.~Kumar, S.~Ermon, and B.~Poole, ``Score-based generative modeling through stochastic differential equations,'' \emph{arXiv preprint arXiv:2011.13456}, 2020.

\bibitem{DDPMSegmentation}
D.~Baranchuk, I.~Rubachev, A.~Voynov, V.~Khrulkov, and A.~Babenko, ``Label-efficient semantic segmentation with diffusion models,'' \emph{arXiv preprint arXiv:2112.03126}, 2021.

\bibitem{DiffusionDet}
S.~Chen, P.~Sun, Y.~Song, and P.~Luo, ``Diffusiondet: Diffusion model for object detection,'' in \emph{Proceedings of the IEEE/CVF international conference on computer vision}, 2023, pp. 19\,830--19\,843.

\bibitem{diff3detr}
J.~Deng, J.~Lu, and T.~Zhang, ``Diff3detr: Agent-based diffusion model for semi-supervised 3d object detection,'' \emph{arXiv preprint arXiv:2408.00286}, 2024.

\bibitem{diffbev}
J.~Zou, K.~Tian, Z.~Zhu, Y.~Ye, and X.~Wang, ``Diffbev: Conditional diffusion model for bird’s eye view perception,'' in \emph{Proceedings of the AAAI Conference on Artificial Intelligence}, vol.~38, no.~7, 2024, pp. 7846--7854.

\bibitem{latent_diffusion}
R.~Rombach, A.~Blattmann, D.~Lorenz, P.~Esser, and B.~Ommer, ``High-resolution image synthesis with latent diffusion models,'' in \emph{Proceedings of the IEEE/CVF conference on computer vision and pattern recognition}, 2022, pp. 10\,684--10\,695.

\bibitem{grap-matching1}
J.~E. Hopcroft and R.~M. Karp, ``An n\^{}5/2 algorithm for maximum matchings in bipartite graphs,'' \emph{SIAM Journal on computing}, vol.~2, no.~4, pp. 225--231, 1973.

\bibitem{KM}
H.~Zhu, M.~Zhou, and R.~Alkins, ``Group role assignment via a kuhn--munkres algorithm-based solution,'' \emph{IEEE Transactions on Systems, Man, and Cybernetics-Part A: Systems and Humans}, vol.~42, no.~3, pp. 739--750, 2011.

\bibitem{levenberg}
K.~Levenberg, ``A method for the solution of certain non-linear problems in least squares,'' \emph{Quarterly of applied mathematics}, vol.~2, no.~2, pp. 164--168, 1944.

\bibitem{relu}
Y.~Chen, X.~Dai, M.~Liu, D.~Chen, L.~Yuan, and Z.~Liu, ``Dynamic relu,'' in \emph{European conference on computer vision}.\hskip 1em plus 0.5em minus 0.4em\relax Springer, 2020, pp. 351--367.

\bibitem{KL}
A.~Malinin and M.~Gales, ``Reverse kl-divergence training of prior networks: Improved uncertainty and adversarial robustness,'' \emph{Advances in neural information processing systems}, vol.~32, 2019.

\bibitem{non}
S.~R. De~Groot and P.~Mazur, \emph{Non-equilibrium thermodynamics}.\hskip 1em plus 0.5em minus 0.4em\relax Courier Corporation, 2013.

\bibitem{markov}
J.~R. Norris, \emph{Markov chains}.\hskip 1em plus 0.5em minus 0.4em\relax Cambridge university press, 1998, no.~2.

\bibitem{unet}
O.~Ronneberger, P.~Fischer, and T.~Brox, ``U-net: Convolutional networks for biomedical image segmentation,'' in \emph{Medical image computing and computer-assisted intervention--MICCAI 2015: 18th international conference, Munich, Germany, October 5-9, 2015, proceedings, part III 18}.\hskip 1em plus 0.5em minus 0.4em\relax Springer, 2015, pp. 234--241.

\bibitem{condition1}
A.~Ramesh, M.~Pavlov, G.~Goh, S.~Gray, C.~Voss, A.~Radford, M.~Chen, and I.~Sutskever, ``Zero-shot text-to-image generation,'' in \emph{International conference on machine learning}.\hskip 1em plus 0.5em minus 0.4em\relax Pmlr, 2021, pp. 8821--8831.

\bibitem{f-cooper}
Q.~Chen, X.~Ma, S.~Tang, J.~Guo, Q.~Yang, and S.~Fu, ``F-cooper: Feature based cooperative perception for autonomous vehicle edge computing system using 3d point clouds,'' in \emph{Proceedings of the 4th ACM/IEEE Symposium on Edge Computing}, 2019, pp. 88--100.

\bibitem{v2vnet}
T.-H. Wang, S.~Manivasagam, M.~Liang, B.~Yang, W.~Zeng, and R.~Urtasun, ``V2vnet: Vehicle-to-vehicle communication for joint perception and prediction,'' in \emph{Computer Vision--ECCV 2020: 16th European Conference, Glasgow, UK, August 23--28, 2020, Proceedings, Part II 16}.\hskip 1em plus 0.5em minus 0.4em\relax Springer, 2020, pp. 605--621.

\bibitem{opencda}
M.~Rosenman and F.~Wang, ``A component agent based open cad system for collaborative design,'' \emph{Automation in Construction}, vol.~10, no.~4, pp. 383--397, 2001.

\bibitem{carla}
A.~Dosovitskiy, G.~Ros, F.~Codevilla, A.~Lopez, and V.~Koltun, ``Carla: An open urban driving simulator,'' in \emph{Conference on robot learning}.\hskip 1em plus 0.5em minus 0.4em\relax PMLR, 2017, pp. 1--16.

\bibitem{DDPM}
J.~Ho, A.~Jain, and P.~Abbeel, ``Denoising diffusion probabilistic models,'' \emph{Advances in neural information processing systems}, vol.~33, pp. 6840--6851, 2020.

\bibitem{DDIM}
J.~Song, C.~Meng, and S.~Ermon, ``Denoising diffusion implicit models,'' \emph{arXiv preprint arXiv:2010.02502}, 2020.

\end{thebibliography}

\newpage

\vspace{11pt}




\vfill

\end{document}